\documentclass[letterpaper, 10 pt, conference]{ieeeconf}  

\IEEEoverridecommandlockouts                              

\usepackage{amsmath} 
\usepackage{xspace}
\usepackage{xcolor}
\usepackage{graphicx}
\usepackage{url}
\usepackage{cite}
\usepackage{soul}

\makeatletter
\DeclareRobustCommand\onedot{\futurelet\@let@token\@onedot}
\def\@onedot{\ifx\@let@token.\else.\null\fi\xspace}

\def\eg{\emph{e.g}\onedot} 
\def\ie{\emph{i.e}\onedot} 
\def\cf{\emph{c.f}\onedot} 
 
\def\wrt{w.r.t\onedot} 

\def\Fig{Fig\onedot}   \def\Tab{Table}  
\makeatother

\newcommand{\PAR}[1]{\vskip4pt \noindent{\bf #1~}}

\newcommand{\bOmega}{\boldsymbol{\Omega}}
\newcommand{\bP}{\mathbf{P}}
\newcommand{\bT}{\mathbf{T}}
\newcommand{\bu}{\mathbf{u}}
\DeclareMathOperator*{\argmin}{argmin~}
\newcommand{\norm}[1]{\left\lVert#1\right\rVert}
\newcommand{\boldparagraph}[1]{\vspace{0.2cm}\noindent{\bf #1:} }

\title{\LARGE \bf
Efficient 2D-3D Matching for Multi-Camera Visual Localization
}

\author{Marcel Geppert$^{1}$, Peidong Liu$^{1}$, Zhaopeng Cui$^{1}$, Marc Pollefeys$^{1,2}$, Torsten Sattler$^{3}$
\thanks{$^{1}$Computer  Vision  and  Geometry  Group,  Department  of  Computer  Science, ETH Z\"urich, Switzerland}
\thanks{$^{2}$Microsoft, Switzerland}%
\thanks{$^{3}$Chalmers University of Technology}%
}

\begin{document}

\maketitle
\thispagestyle{empty}
\pagestyle{empty}

\begin{abstract}
Visual localization, \ie, determining the position and orientation of a vehicle with respect to a map, is a key problem in autonomous driving. 
We present a multi-camera visual inertial localization algorithm for large scale environments. To efficiently and effectively match features against a pre-built global 3D map, we propose a prioritized feature matching scheme for multi-camera systems. In contrast to existing works, designed for monocular cameras, we (1) tailor the prioritization function to the multi-camera setup and (2) run feature matching and pose estimation in parallel. This significantly accelerates the matching and pose estimation stages and allows us to dynamically adapt the matching efforts based on the surrounding environment. In addition, we show how pose priors can be integrated into the localization system to increase efficiency and robustness. Finally, we extend our algorithm by fusing the absolute pose estimates with motion estimates from a multi-camera visual inertial odometry pipeline (VIO). This results in a system that provides reliable and drift-less pose estimation. Extensive experiments show that our localization runs fast and robust under varying conditions, and that our extended algorithm enables reliable real-time pose estimation. 
%
\end{abstract}
\section{INTRODUCTION}


Visual localization is the problem of estimating the position and orientation, \ie, the camera pose, from which a given query image was taken. 
This problem plays a key role in autonomous navigation, \eg, for self-driving cars~\cite{Haene2017IMAVIS} and in Simultaneous Localization and Mapping (SLAM)~\cite{MurAtal2015TRO}. It is also encountered in many 3D computer vision algorithms such as Structure-from-Motion (SfM)~\cite{Schoenberger2016CVPR}, camera calibration~\cite{Haene2017IMAVIS}, and Augmented Reality~\cite{Lynen2015RSS,Middelberg2014ECCV}.

State-of-the-art approaches for visual localization are \emph{structure-based}, \ie, they explicitly or implicitly use a 3D model to represent the scene. 
Explicit methods typically employ a sparse 3D point cloud constructed via SfM~\cite{Li2012ECCV,Lynen2015RSS,Sattler2017PAMI,Svarm2017PAMI,Zeisl2015ICCV}, allowing them to associate each 3D point with one or more local image descriptors. 
For a given query image, they establish a set of 2D-3D correspondences by comparing the descriptors of local features extracted from the image with the 3D point descriptors. 
Using these matches, they then estimate the camera pose of the query by applying an $n$-point-pose solver~\cite{Haralick94IJCV,Kukelova13ICCV,Lee2015IJRR} inside a RANSAC loop~\cite{Fischler81CACM}. 
In contrast, implicit approaches~\cite{Brachmann2018CVPR,Cavallari2017CVPR,Massiceti2017ICRA,Shotton2013CVPR} forego explicit descriptor matching. 
Instead, they directly learn the 2D-3D matching function by learning a mapping from image patches to 3D scene point coordinates. 
Again, the resulting 2D-3D correspondences are used for RANSAC-based pose estimation. 
Implicit approaches can achieve a higher pose accuracy compared to explicit ones~\cite{Cavallari2017CVPR,Brachmann2018CVPR}. 
Yet, they currently do not scale to larger outdoor scenes~\cite{Brachmann2018CVPR,Schoenberger2018CVPR}. 

Most explicit structure-based localization methods focus on the monocular (single image) case, \eg, Augmented Reality on smartphones and tablets~\cite{Arth09ISMAR,Klein07ISMAR,Lynen2015RSS}, by developing strategies for efficient matching~\cite{Li2010ECCV,Sattler2017PAMI} or for scaling to larger or more complex scenes~\cite{Svarm2017PAMI,Liu2017ICCV,Zeisl2015ICCV,Sattler2017CVPR}. 
Yet, many robotics applications, especially self-driving cars~\cite{Haene2017IMAVIS,Schwesinger2016IV}, benefit from using a multi-camera systems that covers the full $360^\circ$ field-of-view (FoV) around the robot. 
It has also been shown that cameras covering a larger FoV can be localized more accurately~\cite{Arth2011ISMAR} and that multi-camera systems significantly boost localization performance in challenging conditions~\cite{Sattler2018CVPR}. 

Existing work on multi-camera localization has mainly focused on stereo SLAM~\cite{Liu2018IROS,Mur2017TRO,Heng2015AR}, camera calibration~\cite{Heng2015AR,Heng2015JFR}, and camera pose estimation~\cite{Camposeco2016ECCV,Lee2015IJRR,Sweeney20163DV,Ventura2014CVPR}. 
The latter two types of approaches model multi-camera systems as a generalized camera~\cite{Pless2003CVPR}, \ie, a camera with multiple centers of projection, to derive (minimal) solvers for pose estimation.  
Yet, one central aspect of multi-camera localization has received little attention: 
Using multiple images leads to more features that need to be considered during feature matching and thus to significantly longer run-times. 
\begin{figure}[t]
	\centering
	\includegraphics[width=0.8\linewidth]{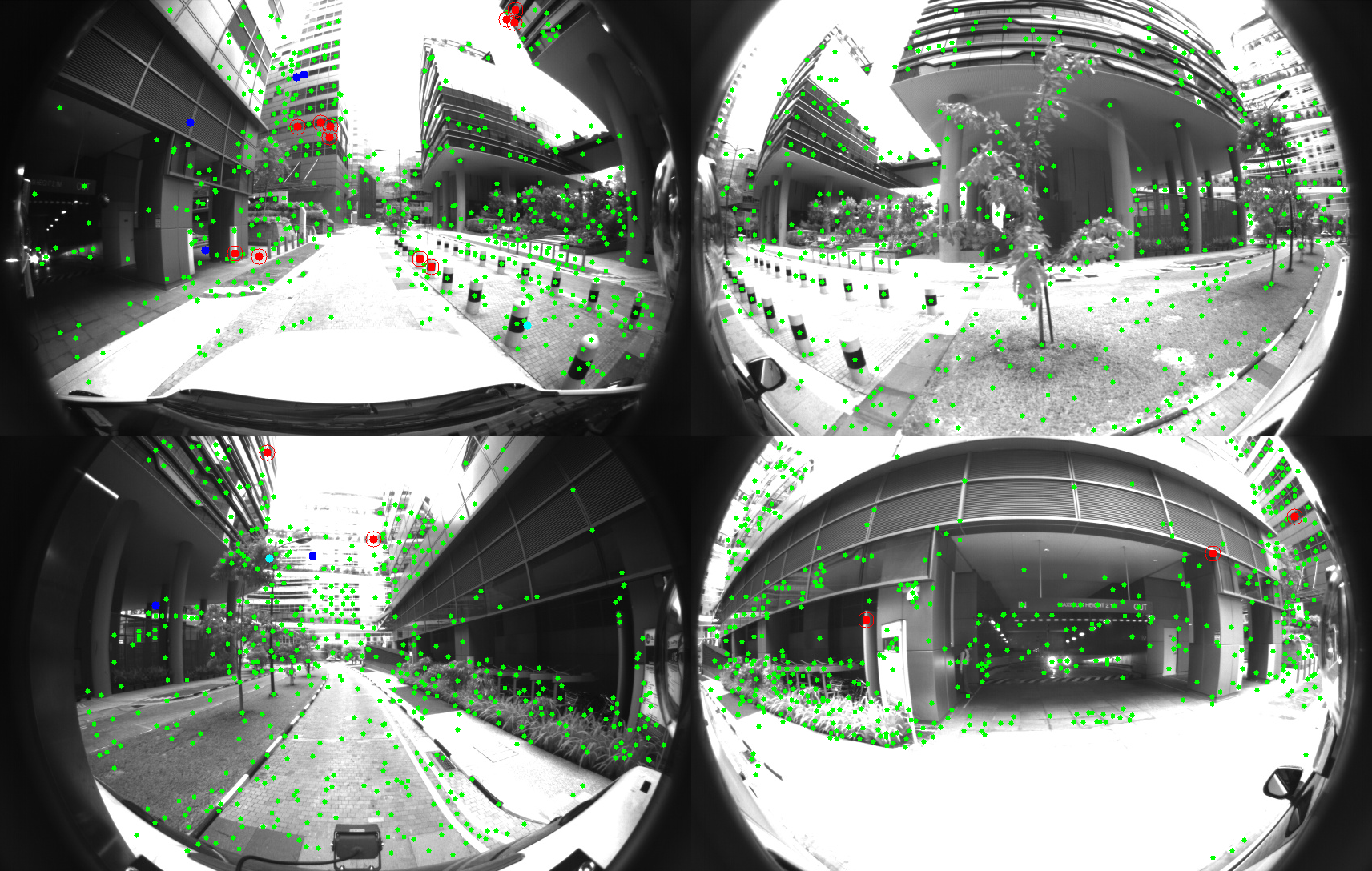}
	\caption{Visual localization based on 2D-3D matching against a 3D scene model for a multi-camera system. We use one fisheye camera mounted on each side of a car. We show the correct matches (red), outlier 2D-3D matches (dark blue), outlier 3D-2D matches (light blue), and unmatched image features (green).}
	\label{fig:one_north_frame_example}
\end{figure}

This paper aims to close this gap in the literature by focusing on efficient 2D-3D matching for multi-camera systems. 
To this end, we make the following main contributions: 
1) We develop a prioritized descriptor matching scheme for multi-camera systems. 
Our strategy is based on Active Search~\cite{Sattler2017PAMI}, an efficient prioritization scheme developed for monocular cameras. 
We show that a fast variant of Active Search, which leads to unstable pose estimates for a single image, is very well suited for multi-camera systems. 
2) We interleave prioritized matching with camera pose estimation. 
In contrast to standard schemes, which terminate search once a fixed number of matches has been found, our approach terminates as soon as sufficiently many geometrically consistent matches have been found. 
3) Inspired by approaches for geometric outlier filtering~\cite{Svarm2017PAMI,Zeisl2015ICCV}, we develop an efficient geometric verification step that can be used to integrate potential pose priors. 
This allows us to avoid comparing descriptors for geometrically implausible matches, which can make our search both more efficient and robust.  
These later two contributions are not restricted to the multi-camera case but also applicable in the monocular scenario. 
4) We show how to combine our approach with a VIO pipeline, enabling our system to provide accurate, drift-free pose estimates in real-time on a car. 

%
\section{Related Work}
\label{sec:related_work}
In the following, we review related work from the areas of visual localization and multi-camera pose estimation. 

\PAR{Efficient visual localization} approaches aim at accelerating the localization process~\cite{Li2010ECCV,Sattler2017PAMI,Brachmann2018CVPR,Cavallari2017CVPR,Massiceti2017ICRA,Kendall2017CVPR,Walch2017ICCV,Valada2018ICRA}. 
Most related to our approach are explicit methods based on prioritized matching~\cite{Li2010ECCV,Sattler2017PAMI}. 
These methods aim at designing a prioritization function such that features that are more likely to yield 2D-3D matches are considered first. 
Once a fixed number of correspondences has been found, matching is terminated and RANSAC-based pose estimation is performed.  
In this paper, we build upon Active Search~\cite{Sattler2017PAMI}. 
We show that a variant of it that is more efficient, but leads to inferior results for monocular images, is actually well-suited for multi-camera systems. 
We adapt the prioritization scheme to encourage distributing matches over many images in the camera system. 
We also propose an adaptive criterion that terminates matching once a certain number of correct matches is found rather than stopping search after finding a fixed number of (possibly incorrect) correspondences.

\PAR{Scalable visual localization.} 
In larger or more complex scenes, which are often characterized by more ambiguous scene elements, it is hard to distinguish between correct and incorrect matches based on descriptor comparisons alone. 
State-of-the-art methods for scalable localization thus relax the matching criteria and perform geometric reasoning to handle the resulting large amounts of wrong matches~\cite{Svarm2017PAMI,Zeisl2015ICCV,Camposeco2017CVPR,Larsson2016BMVC}. 
As a result, they are often too slow for real-time processing. 
In this paper, we propose a geometric filter based on a potentially available pose prior, \eg, from VIO-based camera tracking or via a GPS sensor. 
We show that this filter can be implemented very efficiently, allowing us to perform it before descriptor matching. 
This leads to faster matching times, but also makes matching more robust as we can again relax the matching criteria. 


\PAR{Learning-based methods} integrate machine learning into the localization process. 
This is usually done by either learning the 2D-3D matching stage~\cite{Brachmann2018CVPR,Cavallari2017CVPR,Massiceti2017ICRA} or by directly regressing the camera pose~\cite{Kendall2017CVPR,Walch2017ICCV,Valada2018ICRA}. 
However, recent work shows that these methods are either less accurate than feature-based methods such as the one presented in this paper~\cite{Walch2017ICCV} or do not scale to larger scenes~\cite{Brachmann2018CVPR,Kendall2017CVPR,Schoenberger2018CVPR,Sattler2018CVPR}. 
As such, we do not consider them further in this work.

\PAR{Multi-camera pose estimation algorithms} model multi-camera systems mathematically as a generalized camera~\cite{Pless2003CVPR}. 
Given a set of 2D-3D matches between features in the images of the multi-camera system and 3D scene points, they then estimate the pose of the generalized camera~\cite{Camposeco2016ECCV,Lee2015IJRR,Sweeney20163DV,Ventura2014CVPR}. 
These approaches can easily replace monocular $n$-point-pose solvers inside the RANSAC loop. 
Yet, we are only aware of a single multi-camera localization approach~\cite{Sattler2018CVPR}. 
This method first applies Active Search on all images independently and then performs pose estimation as a post-processing step. 
In contrast, this paper proposes a variant of Active Search that jointly considers all images and we show that this approach leads to significantly faster run-times.

\section{Algorithm}
\begin{figure}
	\centering
	\includegraphics[width=0.95\linewidth]{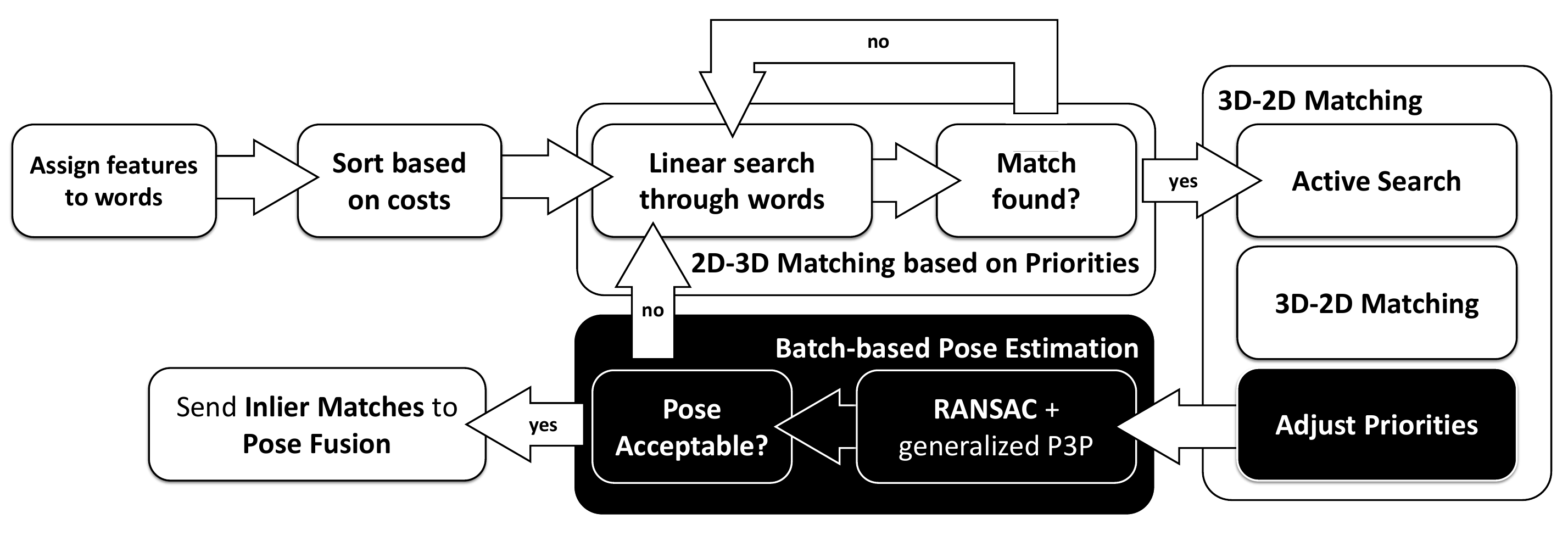}
	\vspace{-0.8em}
	\caption{The localization pipeline. White boxes denote components based on the Active Search approach from~\cite{Sattler2017PAMI}. Black boxes correspond to our modifications made to adapt to the multi-camera system.%
	}
	\vspace{-0.8em}
	\label{fig:localization:pipeline}
\end{figure}

Our system consists of two main functional modules, the localization and pose fusion module.
The localization module provides absolute pose estimates based on a prebuilt map.
It uses two main processes, (1) prioritized feature matching, and (2) iterative RANSAC pose estimation.
An overview over our approach is shown in \Fig~\ref{fig:localization:pipeline}.
Furthermore, the localization module provides a pose prior-based candidate filtering solution that increases its efficiency and robustness when a pose prior is available.
The pose fusion module is used to provide reliable drift-less global pose estimation at camera frame rate.
To this end, we fuse the 2D-3D correspondences from our localization algorithm with the estimated local motions from VIO via a pose graph.
We will describe the details as follows.

\subsection{Map creation}
\label{sec:mapping}
We represent our global map by a sparse 3D point cloud.
Each 3D point from the global map associates with one or multiple 2D feature descriptors, such that 2D-3D feature matching can be performed for visual localization.

We build our map based on the COLMAP SfM pipeline \cite{Schoenberger2016CVPR} by the aid of image synchronized GPS/INS poses.
In particular, since the captured full frame rate images provide redundant information, we subsample a set of representative image frames for 3D reconstruction by using the pose prior provided by GPS/INS. 
A guided 3D reconstruction based on COLMAP is then performed on these selected frames. 

We extract SIFT~\cite{Lowe2004IJCV} features and perform pairwise feature matching among nearby images. The camera poses are initialized with the pose priors from GPS/INS.
This step makes our approach significantly more scalable and robust compared to performing standard SfM from scratch.
We perform multiple iterations of 3D point triangulation, point subsampling, bundle adjustment and outlier filtering.
During bundle adjustment, we fix the extrinsic parameters among the cameras of the multi-camera system.
Subsampling the point cloud before bundle adjustment significantly reduces the optimization complexity and runtime.
After the camera poses are optimized, we perform multiple additional optimization iterations without point subsampling and with fixed image poses to increase the density of the final map.

For point description and retrieval we closely follow the original Active Search approach.
We divide the feature descriptor space into a visual vocabulary, and assign each feature that is associated to a 3D point to its closest word.
We average all descriptors that are assigned to both the same word and the same point, and store this average as a descriptor for the respective point in compressed form by using Product Quantization~\cite{Jegou2011TPAMI}.
Finally, for each point we store which frames observed this point.
Matching an image feature to a point later consists of finding the feature's corresponding word, and then performing linear search through this word's assigned point descriptors.
The vocabulary size is determined empirically so that we achieve a balanced runtime distribution between word assignments and individual feature matching during localization.

For the map size presented in this paper, mapping takes a few hours on a powerful desktop computer.
The map optimization is often limited by pure sequential computations. However, feature extraction and matching make up a significant part of the runtime and can easily be parallelized.

\subsection{Visual localization}
\label{sec:localization}
We design our visual localization algorithm by extending the Active Search approach from \cite{Sattler2017PAMI} for a multi-camera system.
Our algorithm consists of prioritized feature matching and iterative RANSAC pose estimation.
We aim to minimize the number of required feature matches and dynamically adapt the matching efforts to the surrounding environment.
Therefore we run both components in parallel, and add new feature matches iteratively to the pose estimation until a valid pose is found or no more image features are available.

\boldparagraph{Prioritized Feature Matching for Multi-Camera-Systems}
Using all features that can be found in a whole frame usually leads to far more correspondences than required to reliably localize the frame.
In an ideal case, three 2D-3D correspondences are sufficient to compute the pose.
However, due to the presence of outliers a larger number of constraints is required to verify the estimated pose.
Still, this number is far below the total number of features in a frame, and can be reduced further if a subset of correspondences with higher inlier ratio than the whole set is selected.

Active Search implements an ordered feature matching strategy that aims to prioritize more unique, and therefore simpler to match features.
To this end, each feature is first assigned to a word in the map's visual vocabulary using fast nearest neighbor search~\cite{Malkov2016ArXiv}.
The matching order is then determined by sorting the features \wrt to the number of 3D point descriptors assigned to the same word.
A lower number of assigned descriptors suggests that the feature is more unique and has a higher likelihood of being matched correctly.
As a side effect, performing linear search is faster with fewer descriptors, even if no match can be found.

Using feature matches with different viewing directions avoids errors due to quasi-degenerate correspondence sets, and generally leads to more stable pose estimates. As we concentrate on a multi-camera setup, there is an easy way to enforce a more balanced spatial feature distribution by enforcing a balanced feature distribution over the different cameras. This obviously only holds under our current assumption that the cameras have no overlapping fields of view. We add an additional matching cost factor
\begin{equation}
\vspace{-0.4em}
c_I = \frac{\log{m_I + 1}}{\log{6}} + 1 \label{eq:active_search:weighting}
\end{equation}
to direct the prioritization algorithm to preferring a more balanced feature distribution.
For each image $I$, the cost factor $c_I$ increases with the number of matched features $m_I$.
The term grows rapidly in the beginning, leading the prioritized search to consider other images.
Once a few matches are found in all images, the prioritization scheme starts to converge to a situation where all images are (more or less) treated similarly.
The feature order for each single image is independent of this term.
Therefore we can maintain the order for each image separately, and simply choose the next feature based on the scaled cost of the respective image's next feature.

After selecting a feature for matching, we find the two most similar point descriptors within the same word. Here we ignore the possibility of the best match being associated with another word for the sake of simplicity and runtime efficiency.
We reject ambiguous correspondences by performing a bi-directional ratio test~\cite{Lowe2004IJCV}.
First, we threshold the ratio of descriptor distances between the image descriptor and the two point descriptors as in original Active Search.
Afterwards we match the best point descriptor back to the image, and perform a ratio test with the obtained descriptors.
We also reject the match if the generating feature is not the best match for the point.
As in Active Search, we use the newly established match to find 3D neighbors of the point, that were observed together during mapping.
For each point, we match the point descriptor back to the image's feature descriptors as for the ratio test before, and filter the found matches again by a ratio test.
As additional constraint, we also threshold the descriptor distance to the best match to be no more than twice the distance of the generating 2D-to-3D match.
For the monocular camera case, \cite{Sattler2017PAMI} shows that performing this backmatching immediately can lead to correspondence clusters and therefore degenerate or unstable configurations for pose estimation.
However, this is not a concern for our system as we explicitly enforce a sufficient feature distribution in the pose estimation.

\boldparagraph{Iterative RANSAC}
\label{sec:localization:parallel_ransac}
For efficient and robust estimation, we utilize an iterative RANSAC strategy for pose estimation. Instead of attempting to match all descriptors first, and then perform RANSAC on the matches, we run both matching and pose estimation with RANSAC in parallel.
This way we can avoid matching more features than necessary.
If very few matches are sufficient to find a good pose estimate, we can stop matching as soon as this estimate is found.
We utilize Faiss~\cite{Johnson2017ArXiv} and repeatedly match small batches of features to exploit its parallel search capabilities.
After each batch the found matches are immediately handed to the pose estimation algorithm.
While running RANSAC we maintain the best five hypotheses to avoid the need to repeatedly sample the same configurations during the iterative addition of matches.
It is highly likely that a good hypothesis is found very early, but we might not be able to verify it immediately because we do not have enough matches so far.
When the number of inliers increases with the number of matches, we gain confidence that the hypothesis is indeed correct.
It has been shown that utilizing co-occurence information can significantly improve large scale localization \cite{Li2012ECCV}.
To increase the chance to find a correct hypothesis, we guide RANSAC to prefer new and consistent data.
The first match is sampled randomly from the set of recently added matches.
The two other matches are sampled from all available matches so that all 3 points were co-visible during mapping.
Once all recently added matches were sampled multiple times, we switch to a more general method where the first match is sampled randomly from all available matches.
This helps to still find a good hypothesis even if it was missed previously.
When new matched points are obtained, we will first check the best hypotheses.
If the new matches increase the best inlier count and -ratio above the threshold $\eta$, we will stop; otherwise we perform RANSAC as described above and update the best hypotheses if necessary.
To accept a hypothesis, three requirements need to be fulfilled:
The hypothesis has an inlier ratio of at least 20\%, a total number of at least 15 inliers, and the inliers are distributed over more than half of the available cameras.
A match is counted as an inlier for the hypothesis if its angular error is less than 10 degrees.
As soon as a hypothesis is accepted, the parallel feature matching is stopped as well.

\begin{figure}
	\centering
	\includegraphics[width=0.45\linewidth]{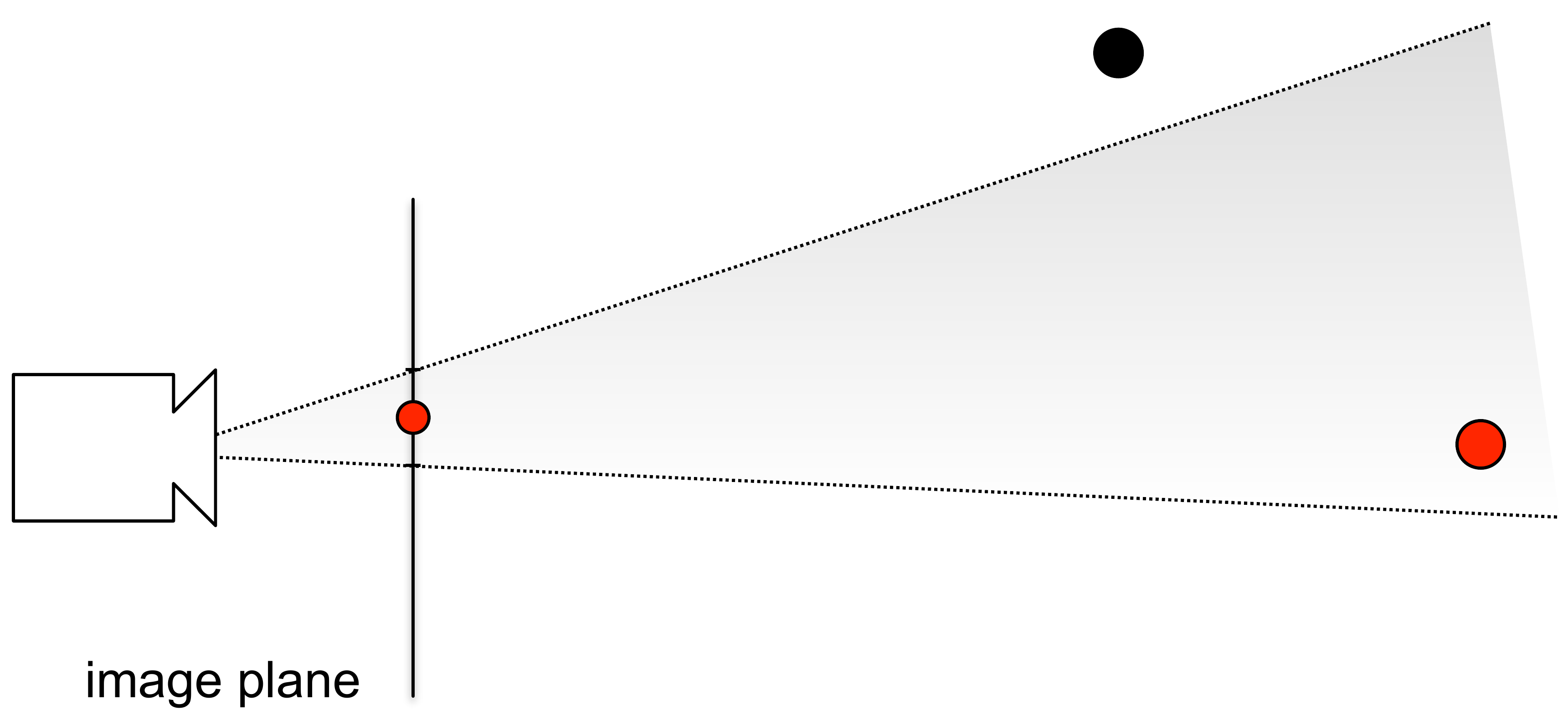}
	\includegraphics[width=0.48\linewidth]{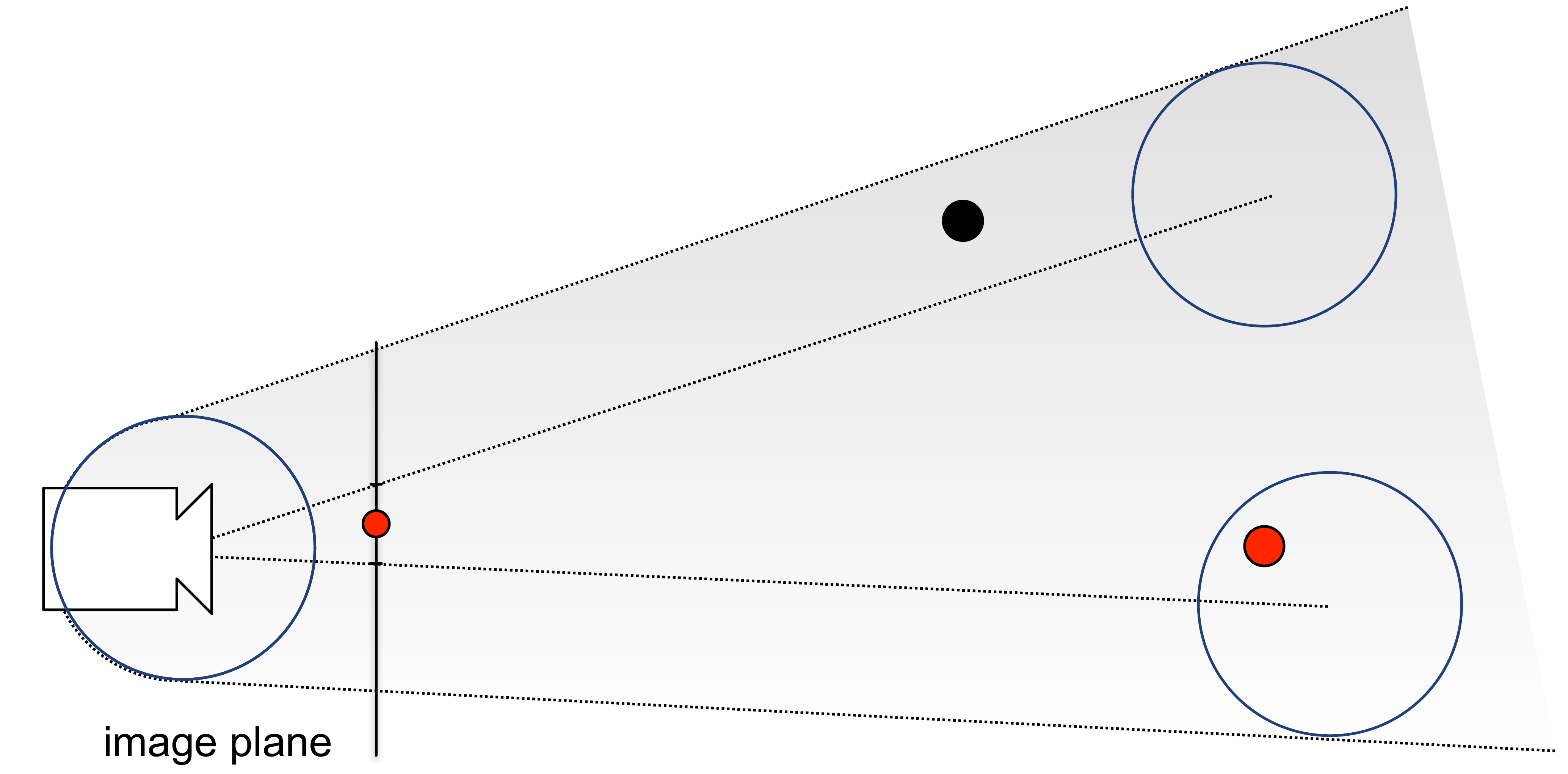}
	\caption{Using pose priors in the localization pipeline: (left) Assuming that the pose of the camera is known precisely, a 3D point (red) is an inlier to the pose if it projects within $r$ pixels of its corresponding image measurement (red point on image plane). This is equivalent to the point lying inside a 3D cone defined by the image measurement and the inlier threshold $r$. 
		(right) Uncertainty on the camera pose can be incorporated by enlarging the cone accordingly. The figure shows the case that the position of the camera is known up to a radius (shown by the circle around the camera). The case of uncertainty in the camera orientation works similarly.}
	\label{fig:localization:priors}
\end{figure}

\boldparagraph{Candidate Filtering with Pose Prior}
\label{sec:localization:pose_prior}
When a pose prior is available during localization in a large map, the number of match candidates drops significantly since we know that many points cannot be seen.
Removing these impossible candidates already before matching limits the necessary effort for descriptor comparisons, and can remove ambiguities from visually similar points in other parts of the map.
This improves both efficiency and robustness of the feature matching.
Assuming the camera pose is known accurately, a 3D point $\mathbf{p}$ forms an inlier match with a feature $f$ if it is projected within $r$ pixels around $f$ in the image.
Equivalently, $\mathbf{p}$ has to lie in a cone along the feature direction whose opening angle $\alpha$ is defined by $r$.
This is illustrated in 2D in \Fig~\ref{fig:localization:priors}.
We slightly adapt this formulation to account for the uncertainty in the pose prior.
For simplicity, we make 2 assumptions on this uncertainty:
1) The real camera position lies within a distance $d$ around the prior position.
2) The real heading direction lies within a heading cone of apex angle $2\theta$ around the prior heading direction.
The heading uncertainty can trivially be incorporated by increasing the cone opening angle
\begin{equation}
\vspace{-0.4em}
\alpha^\prime = \alpha + 2\theta \enspace .
\vspace{-0.4em}
\end{equation}
The camera position uncertainty can be translated into an uncertainty of the 3D point positions.
We can assume a sphere of radius $d$ around each point, and check if this sphere intersects with the cone.
A problem that arises with this approach is that it changes the density of the descriptor space by filtering spatially unreasonable points.
While we can assume that the best match is not filtered out (otherwise it would be a wrong match anyway), it is possible that the next best matches are removed by the spatial constraint.
This impacts the result of the ratio test when performed only on the remaining descriptors.
In contrast, the density of the descriptor space defined by the image features is not affected by the candidate filter.
This makes the backward ratio test in this situation especially useful.
It should also be noted that this filtering approach introduces a hard limit on the uncertainty that can be handled.
If the prior pose error is outside our thresholds, the correct matches cannot be found any more.
For our experiments we therefore use a large but fixed uncertainty of 50 m radius around the given position, and 10$^\circ$ around the given heading.
These values could, however, dynamically be adapted if the prior pose uncertainty can be estimated reliably.

\subsection{Pose fusion with VIO}
The localization algorithm occasionally still outputs wrong pose estimates due to unreliable feature matches in challenging conditions.
Furthermore, localization can only run at a low frequency up to 4Hz due to its computational complexity and feature extraction runtime. In contrast, VIO is able to estimate accurate local vehicle motions at image frame rate ($\sim$30 Hz).
Therefore, we combine our localizer with a multi-camera VIO pipeline~\cite{Liu2018IROS}.

We model the problem of fusing the global poses constraints from our localizer and the estimated trajectories from the sliding window VIO algorithm via a factor graph.
In particular, we create a pose node for every pose estimated by the localizer and add constraints from the estimated 2D-3D matches.
Between each pair of two consecutive pose nodes we add relative pose constraint obtained by integrating VIO poses for the respective time interval.
The factor graph is optimized by minimizing following cost function
\begin{align}
\tiny
\vspace{-0.8em}
&\hat{\bT}_{i}, \hat{\bT}_{i+1},..., \hat{\bT}_{i+N} = \argmin_{\bT_{i}, \bT_{i+1},..., \bT_{i+N}} \\
&\sum_{j=i}^{i+N-1}\norm{\bT_{j+1}^{-1} * \bT_{j} * \Delta{\bT}_{j}^{j+1}}_{\Sigma_{0}}^2 + \\
&\sum_{j=i}^{i+N-1} \sum_{k\in\bOmega_{j}} \norm{\bu_k - \pi(\bP_k, \bT_j)}_{\Sigma_1}^2 \enspace .
\vspace{-0.8em}
\end{align}

Here, $N$ is the sliding window size of the factor graph, $\hat{\bT}_{i}, \hat{\bT}_{i+1},..., \hat{\bT}_{i+N}$ are the estimated global vehicle poses, $\Delta{\bT}_{j}^{j+1}$ is the relative pose constraint from VIO between pose node $j$ and $j+1$, $\Sigma_{0}$ is the covariance matrix for the relative pose, $\bOmega_{j}$ is a set of 2D-3D feature matches found by localization for the $j^{th}$ vehicle pose node, and $\bu_k$ and $\bP_k$ are the 2D and 3D feature measurements of $k^{th}$ correspondence, respectively. 
$\Sigma_1$ is the covariance matrix for re-projection errors, and $\pi(*)$ is the projection function of the camera corresponding to $\bu_k$.
The pose graph is implemented and optimized in the Google Ceres framework~\cite{ceres-solver}.

The above cost function will only be optimized once the visual localizer provides new measurements.
Image frame rate vehicle poses are computed by integrating relative poses estimated by VIO with respect to the latest vehicle global pose node of the factor graph.

\section{Experiments}
\subsection{Dataset selections}
We evaluate our algorithm on two datasets, \emph{One North} and \emph{RobotCar Seasons}.
The One North dataset was recorded in Singapore's One North district as part of the AutoVision project \cite{Heng2019ICRA}.
The dataset contains a traversal of a 5 km trajectory at different times.
We use 3 sets of synchronized frames to create the map and evaluate the localization offline, and a live data stream to evaluate the pose fusion with VIO.
The images of the map creation show strong sunlight and hard shadows that influence the appearance of the scene.
The `sunny' traversal was recorded directly after the mapping set and therefore shows very similar conditions.
The `overcast' traversal, instead, was captured under a significantly different illumination condition.
The live data stream was also recorded during the `overcast' traversal.
We observed that the ground truth for the One North dataset, which is inferred from GPS and IMU, is in itself inconsistent, and shows deviations compared to the map or when closing a loop.
This has an impact on the errors reported in the quantitative analysis.
The RobotCar Seasons benchmark dataset~\cite{Sattler2018CVPR} is a subset of the RobotCar dataset~\cite{Maddern2016IJRR} which was collected in Oxford under changing weather conditions over the course of one and a half years.
The evaluated conditions include, among others, strong sunlight, rain, and snow.
As the ground truth for this benchmark is withheld from the public, we evaluate the localization only without using a pose prior.
The benchmark provides already extracted image features.
For the sake of comparability we do not extract features on our own, but use the provided ones.
However, it should be mentioned that the algorithms parameters were optimized with a few hundred features per image in mind, while the benchmark provides several thousand features per image.

\subsection{Evaluation metrics}
We are interested in two characteristics of the system, accuracy and runtime.
To evaluate the accuracy, the RobotCar Seasons evaluation defines multiple error classes expressed as heading and position error thresholds, and reports the percentage of pose estimates with errors below the respective thresholds.
We adopt this approach for our One North evaluation as well.
However, as mentioned above, the ground truth poses contain errors, most prominently in the altitude.
We therefore limit our position error evaluation to the ground plane and ignore potential altitude errors.
For the runtime analysis, we ignore the time required for feature extraction from the images, as this is not part of our contribution, and only report the time from starting the feature matching until a pose is found or the system reports that it cannot localize, respectively.

\subsection{Parameter settings}
For the One North evaluation we use a camera setup of four wide-angle cameras, one on each side of the vehicle platform \cite{Heng2019ICRA}.
The cameras are rigidly connected and the cameras' intrinsic and extrinsic parameters are known.
For the offline localization, we evaluate a frame after every meter driven.
When evaluating the pose prior, we add random noise to the ground truth to obtain a prior as it could also be observed from a low quality GPS system.
We sample a heading deviation $e_H \sim \mathcal{N}\left(0, 5\right)$ degrees and rotate around the vertical axis.
We also sample an offset distance $e_P \sim \mathcal{N}\left(0,10\right)$ meters and shift the pose in a (uniform) random direction in the horizontal plane.
For the RobotCar benchmark we use the three cameras also used in~\cite{Sattler2018CVPR}, located at the left, right and rear of the vehicle.

\subsection{Experimental evaluations}

\boldparagraph{Localization accuracy}
Localization with the `sunny' traversal can find a pose for more than 97\% of all query frames.
When employing the pose prior, this number increases further, but at the same time the amount of noise in the estimated poses increases as well.

\begin{figure}[t]
	\centering
	\includegraphics[width=0.48\linewidth]{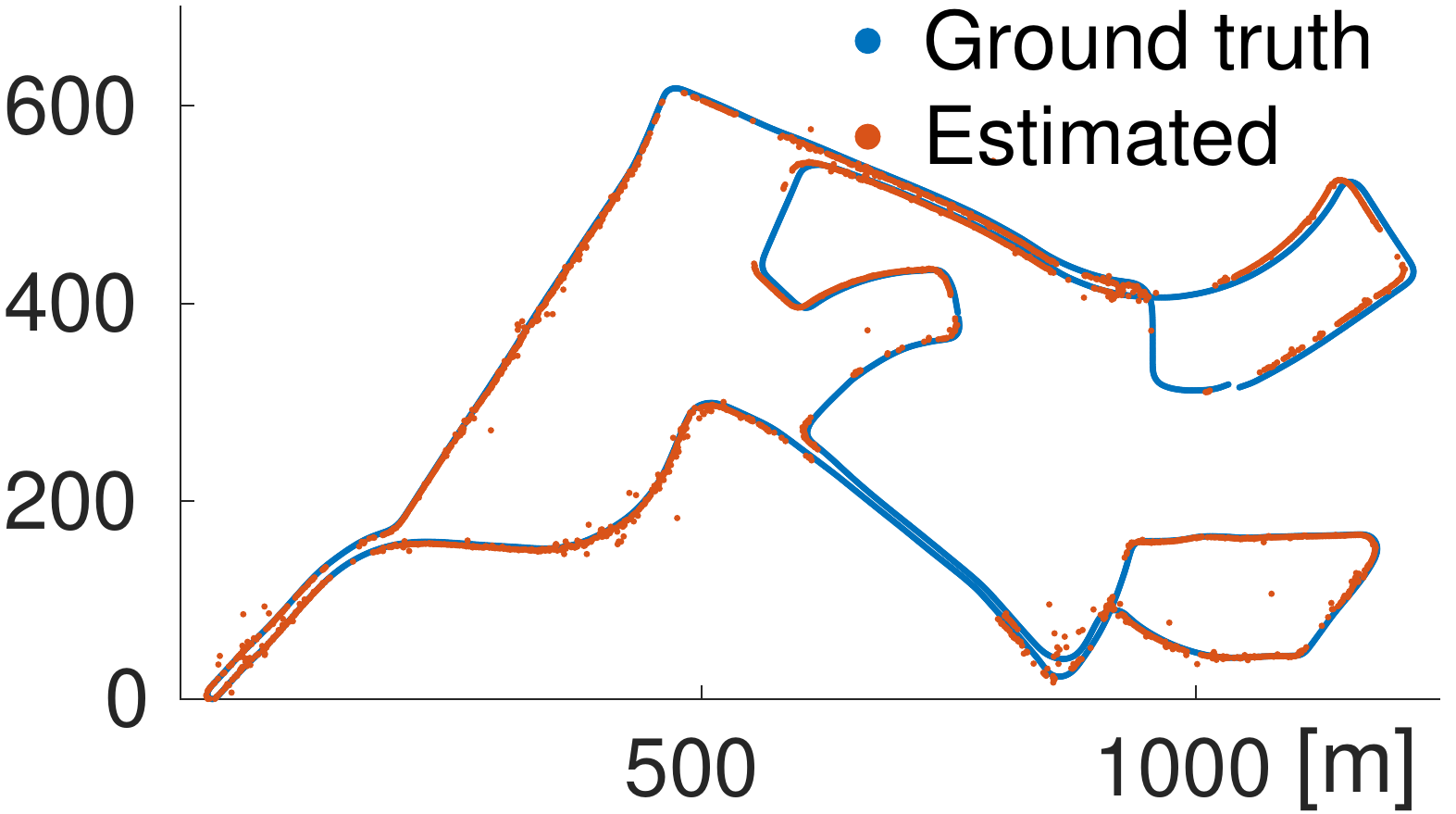}
	\includegraphics[width=0.48\linewidth]{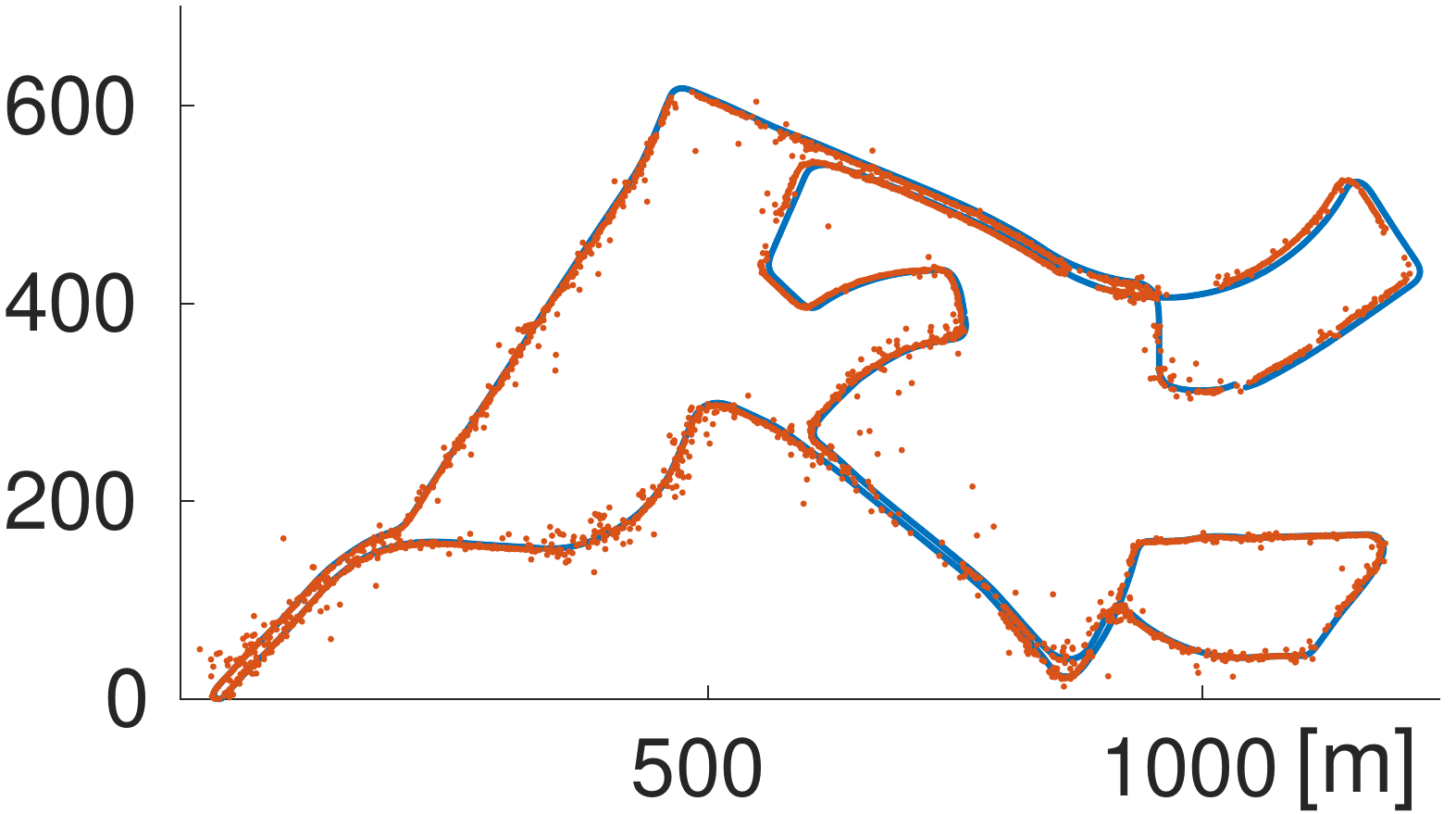}
	\vspace{-1.0em}
	\caption{Estimated vs. ground truth poses for the 'overcast' trajectory without (left) and with (right) a pose prior.}
	\vspace{-1.0em}
	\label{fig:experiments:one_north:gt:trajectory_run1}
\end{figure}

A quantitative analysis is provided in \Tab~\ref{tab:experiments:one_north}.
Noticeable is the almost complete absence of poses with very low errors, \ie, less then 5 degree heading / 0.5 m position deviation.
We suspect this to be caused by misalignments of the map and ground truth.
\begin{table}
	\centering
	\caption{Percentages of poses within different error thresholds and runtimes for the One North datasets. Runtimes are reported excluding feature extraction.}
	\vspace{-1.0em}
	\label{tab:experiments:one_north}
	\begin{tiny}
		\begin{tabular}{l|r|r|r|r|r|r}
			& \multicolumn{5}{c|}{\% of poses within error thresholds (deg / m)} & \multicolumn{1}{c}{Mean runtimes [ms]}\\
			& 2 / 0.25 & 5 / 0.5 & 10 / 5 & 15 / 10 & 20 / 20 & \\ \hline
			sunny 			& 0.1 & 1.8 & 97.2 & 97.8 & 98.0 & 45\\ \hline
			sunny+prior 	& 0.3 & 1.6 & 95.9 & 98.2 & 99.0 & 40\\ \hline
			overcast 		& 0.0 & 0.1 & 39.5 & 51.7 & 60.7 & 111 \\ \hline
			overcast+prior  & 0.0 & 0.1 & 43.6 & 63.0 & 75.6 & 87 \\ \hline			
		\end{tabular}
	\end{tiny}	
	\vspace{-0.4em}
\end{table}
For the `overcast' traversal, the localization accuracy significantly decreases, and in some sections of the trajectory localization fails consistently.
Using the pose prior again increases the percentage of localized poses, but also noise as shown in \Fig~\ref{fig:experiments:one_north:gt:trajectory_run1}. 
Notice that the failure to localize every frame as well as inaccurate poses can be handled by integrating localization into a VIO system, as shown below.
\begin{figure}
	\centering
	\includegraphics[width=0.7\linewidth]{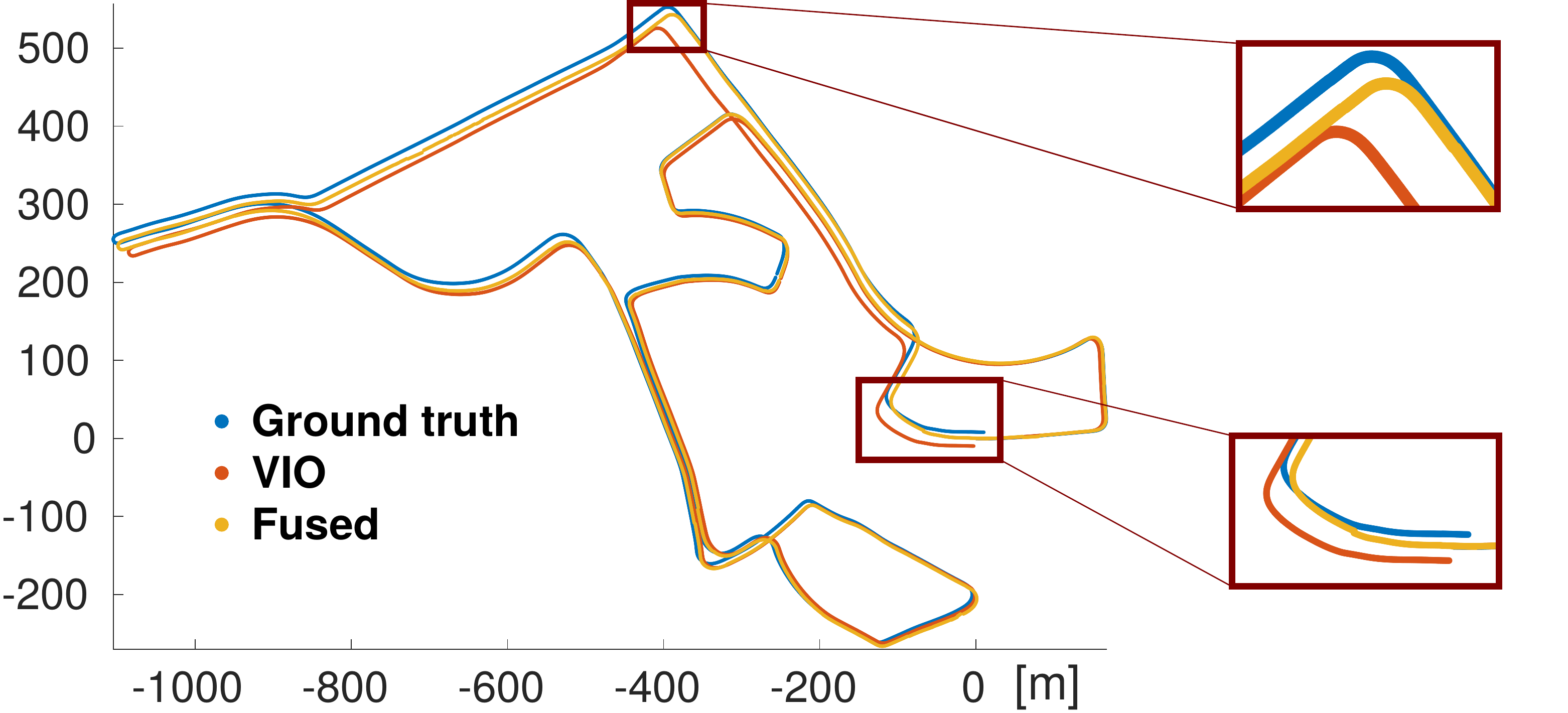}
	\vspace{-1.0em}
	\caption{Trajectory estimated by pure VIO and fusion of VIO and localization from the `overcast' dataset vs. ground truth. The fused pose actually reaches the starting point, whereas there is an offset in the GPS ground truth.}
	\label{fig:experiments:one_north:trajectory_run1_fused}
\end{figure}

We additionally evaluated the algorithm with different camera setups on the `overcast' dataset.
Interestingly, the accuracy seems to be much more dependent on the viewing directions than on the number of used cameras.
While the front facing camera seems to be the most useful, the back facing camera only provides very little benefit.
For the RobotCar Seasons benchmark, we compare our method to both the original monocular Active Search algorithm, as well as a version that performs Active Search on each image individually and then performs multi-camera pose estimation (Active Search + GC). 
With our default parameters, our approach performs significantly worse than both Active Search references for low error classes, but the difference shrinks with higher error thresholds.
We found our default parameters for the feature matching ratio test to be not strict enough for the much more dense map of the RobotCar dataset.
After replacing our thresholds with the one used by Active Search our accuracy ranges between the the monocular and generalized Active Search algorithm.
Interestingly, these changes decrease the accuracy for night time localization.
The results are shown in  \Tab~\ref{tab:experiments:robotcar:summarized_scenes}.
\begin{table}[t]
	\centering
	\caption{Summarized percentages of poses within error thresholds for all day and night scenes in RobotCar Seasons with different setups.}
	\vspace{-1.0em}
	\label{tab:experiments:robotcar:summarized_scenes}
	\begin{tiny}
		\begin{tabular}{l|c|c}
			& \textbf{all day} & \textbf{all night} \\
			\multicolumn{1}{r|}{deg}	& 2 / 5 / 10 & 2 / 5 / 10 \\
			\multicolumn{1}{r|}{m}		& .25 / .50 / 5.0 & .25 / .50 / 5.0\\ \hline
			Active Search~\cite{Sattler2017PAMI}	& 35.6 / 67.9 / 90.4 & 0.9 / 2.1 / 4.3 \\ \hline
			Active Search + GC~\cite{Sattler2018CVPR} & 45.5 / 77.0 / 94.7 & 2.7 / 6.9 / 12.1 \\ \hline
			Ours (GC)	& 15.5 / 40.5 / 85.8 & 0.3 / 1.9 / 8.2 \\ \hline
			Ours (GC) + Active Search matching thresholds & 41.6 / 73.3 / 90.1 & 0.2 / 1.1 / 2.6 \\ \hline
			
		\end{tabular}
	\end{tiny}	
	\vspace{-0.3em}
\end{table}

\boldparagraph{Localization runtime}
Due to the dynamic number of matches, our algorithm's runtimes are highly dependent on the recognizability of the scene.
This can be seen in \Tab~\ref{tab:experiments:one_north}.

The `sunny' traversal can easily be located within our map with very few good feature matches, which leads to very low average runtimes of 48ms and 55ms with and without prior, respectively.
For the `overcast' images, more or even all image features need to be matched, resulting in significantly higher runtimes.
In this case, using the pose prior leads to significantly lower runtimes, whereas its impact for `sunny' traversal, where very few features need to be matched anyway, is mostly negligible.
Similarly, the runtime scales less than linear with the number of cameras since the number of matched features is independent of the number of input images. Therefore, limiting the available field of view to reduce the runtime generally seems not worthwhile.

Analyzing the runtimes for the RobotCar Seasons experiments shows that our approach with a multi-camera setup easily outperforms the Active Search multi-camera implementation and is only slightly slower than the monocular camera version.
Active Search and multi-camera Active Search report runtimes of 291ms and 879ms on average, while our solution takes 371ms.
However, the average runtimes for our system are significantly influenced by the frames that could not be localized, as our system tries to match all available image features in this case.
This could easily be avoided by setting a (large) upper limit on the number of matched features.
If we only consider localized frames for our solution, our system is, with 239ms on average for three cameras, almost 20\% faster than reported for the monocular Active Search implementation.

\boldparagraph{Pose fusion evaluation}
Fusing the localizer's absolute pose estimates with relative pose estimates from VIO enables us to provide the current pose at a much higher frame rate and trajectory smoothness.
\Fig~\ref{fig:experiments:one_north:trajectory_run1_fused} shows the positions estimated by pure VIO on the `overcast' traversal of the OneNorth dataset, which experiences drift over the course of the trajectory, together with the pose fusion result.
Here, the drift can successfully be corrected using localization against the map.
Integrating localization results into the VIO pipeline also allows us to track the pose over the full sequences, even if localization fails or is inaccurate in some parts (\cf~\Fig~\ref{fig:experiments:one_north:gt:trajectory_run1}).

\section{Conclusion}
In this paper, we have proposed a novel visual localization system tailored to multi-camera setups. 
Our approach does not require preset parameters for the number of features to match, and is therefore highly adaptive to different scenes while minimizing the feature matching effort.
We also presented an efficient candidate filtering step based on pose priors.
By fusing our pose estimates with relative motion estimates from a VIO system we are able to provide accurate absolute poses in real time.

In the future, we plan to optimize parameters and our implementation to leverage the full potential of the approach. 
Adjusting the pose prior to the uncertainty of the VIO pipeline, rather than using a fixed large uncertainty radius, seems especially promising.

\bibliographystyle{ieee}
\bibliography{egbib,bibliography_marcel}

\end{document}